\documentclass[conference]{IEEEtran}
\IEEEoverridecommandlockouts   
\usepackage{times}

\usepackage[numbers]{natbib}
\usepackage{multicol}

\usepackage{graphicx}
\usepackage{amsmath}
\usepackage{amssymb}
\usepackage{booktabs}
\usepackage{multirow}
\usepackage{xcolor}
\usepackage{url}
\usepackage{subcaption}

\usepackage[bookmarks=true]{hyperref}
\hypersetup{
  pdftitle={Freeze, Share, Shrink: Rethinking the Action Backbone in Diffusion Policies},
  pdfauthor={Jian Zhou, Sihao Lin, Shuai Fu, Zerui Li, Gengze Zhou, Qi Wu}
}

\begin{document}

\title{Freeze, Share, Shrink: Rethinking the\\ Action Backbone in Diffusion Policies}

\author{Jian Zhou$^{1}$, Sihao Lin$^{1}$, Shuai Fu$^{1}$, Zerui Li$^{1}$, Gengze Zhou$^{1}$, Qi Wu$^{1}$%
\thanks{$^{1}$Jian Zhou, Sihao Lin, Shuai Fu, Zerui Li, Gengze Zhou and Qi Wu are with the
Australian Institute for Machine Learning, University of Adelaide,
SA, Australia. {\tt\small \{j.zhou, sihao.lin, shuai.fu,
zerui.li, gengze.zhou, qi.wu01\}@adelaide.edu.au}}%
}

\maketitle

\begin{abstract}
Many recent Vision-Language-Action models employ diffusion or flow-matching backbones with hundreds of millions of parameters for action generation.
However, unlike image synthesis where the output spans millions of diverse pixels, a manipulation policy generates only short sequences of low-dimensional, physically correlated action values, a far simpler target that may not require such capacity.
We confirm this intuition and show that, in modulation-conditioned diffusion policies, task adaptation can be routed entirely through the conditioning pathway, leaving a frozen, observation-free backbone that serves as a reusable trajectory prior.
To establish this, we introduce a decoupled training recipe: a general-purpose action head is first pretrained on observation-free forward-kinematics data, then frozen while only the conditioning pathway is trained for downstream tasks.
Using Diffusion Policy as a testbed, we show that on both MimicGen and LIBERO, a single frozen backbone shared across all tasks matches normally trained counterparts.
In our ablations, this succeeds with modulation-based conditioning, while attention-based conditioning is embedded in the backbone's own weights and collapses once they are frozen.
Ablations show the pretraining signal (joint positions, end-effector poses, or none) has little effect, while a randomly initialized backbone fails entirely, indicating that pretraining need only supply a general trajectory prior.
Finally, a 5M-parameter MLP backbone matches or exceeds both the 244M U-Net and the transformer on these benchmarks, suggesting the action backbone is over-parameterized and that architectures inherited from image and language generation are a poor inductive-bias fit for the low-dimensional action target, with implications for action-backbone design in VLA models and other policies with an action backbone.
\end{abstract}

\IEEEpeerreviewmaketitle

\section{INTRODUCTION}

\begin{figure}[t]
\centering
\includegraphics[width=\columnwidth]{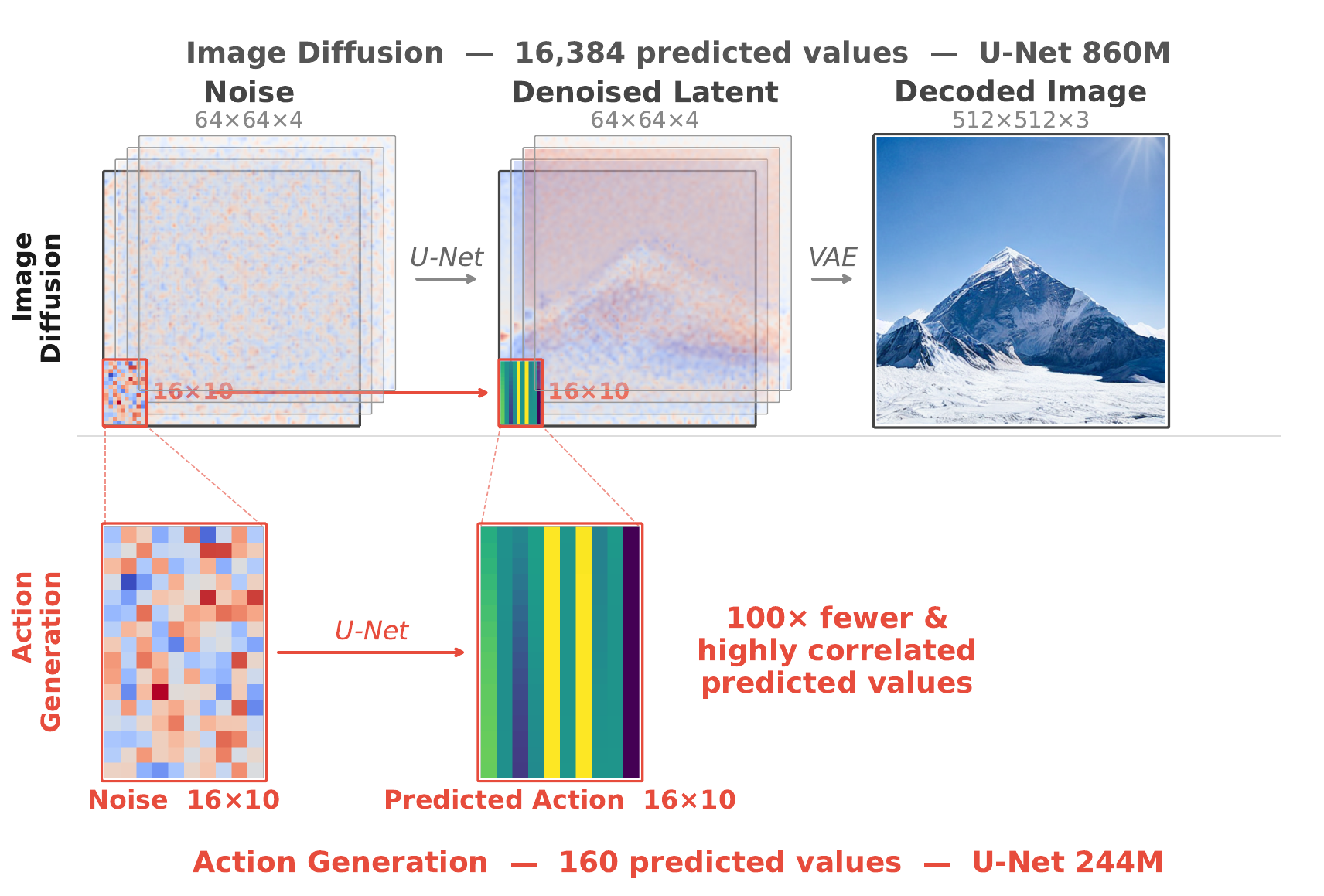}
\caption{\footnotesize Output dimensionality contrast. Image diffusion models such as Stable Diffusion~\cite{stable_diffusion} denoise 16,384 structured latent values, while a manipulation policy generates only 160 physically correlated values. Yet action experts in current VLA designs often adopt similar-scale architectures. Is such capacity necessary, and must the action backbone be trained on task data?}
\label{fig:teaser}
\end{figure}

Diffusion and flow-matching models have become one of the dominant approaches for action generation in robotic manipulation, powering both standalone Diffusion Policies~\cite{DP} and the action experts within Vision-Language-Action (VLA) models such as $\pi_0$~\cite{pi0}, CogACT~\cite{CogACT}, GR00T-N1~\cite{GR00T-N1}, and DexVLA~\cite{DexVLA}.
These action experts borrow architectures from image and video generation (U-Nets, DiT blocks, flow-matching networks) along with their scale: 244M parameters for Diffusion Policy's U-Net, ${\sim}$300M for $\pi_0$'s flow-matching network, and over 1B for DexVLA (Table~\ref{tab:action_experts}).
Yet the output spaces are fundamentally different (Fig.~\ref{fig:teaser}).
Image diffusion models denoise tens of thousands of structured latent values, while a manipulation policy generates only $16 \times 10 = 160$ action values in Diffusion Policy~\cite{DP}, of which only a fraction are executed before replanning (8 steps in DP, 16--25 in $\pi_0$~\cite{pi0}).
This orders-of-magnitude mismatch in output complexity raises a natural question: do these large action backbones actually need their capacity, or is much of it unnecessary?
The question is not merely academic: recent work shows that large uninitialized action experts can destabilize VLA training through harmful gradient flows to the reasoning backbone~\cite{driess2025knowledge}, suggesting that oversized action heads are not just wasteful but actively problematic.

The problem is compounded by data scarcity: robot demonstration data is orders of magnitude smaller than what drives image or language models, making it critical to identify which policy components genuinely require task-specific demonstrations and which do not.

We investigate this through the \textbf{Decoupled Action Expert}, a two-stage training recipe that isolates the role of the action backbone:
\begin{itemize}
    \item \textbf{Stage~1 (Observation-Free Pretraining):} A general-purpose action head is pretrained on kinematic pairs (joint positions mapped to end-effector poses via forward kinematics), available in any trajectory dataset at negligible cost, with no observation-action pairing required.
    \item \textbf{Stage~2 (Task-Specific Adaptation):} The pretrained action head is frozen, and only the newly initialized conditioning pathway is trained on downstream demonstrations.
\end{itemize}
The logic is straightforward: if freezing the backbone preserves task performance, then task adaptation can be carried entirely by the conditioning pathway, and the backbone needs no task-specific training.
Because Stage~1 requires only kinematic trajectories, the recipe can also leverage data sources that standard observation-action pipelines cannot use at all.

Since the action expert in VLA models, such as $\pi_0$'s flow-matching network~\cite{pi0} and CogACT's DiT~\cite{CogACT}, serves the same functional role as the denoising backbone in Diffusion Policy, we use Diffusion Policy as a controlled testbed, free from confounding factors introduced by billion-parameter VLM reasoning components and tractable enough for the extensive controlled experiments across MimicGen~\cite{mimicgen} and LIBERO~\cite{liu2023libero} that this study requires. Our claims therefore concern modulation-conditioned diffusion policies; extending them to full VLA systems is motivated future work (Section~\ref{sec:conclusion}).
Our main contributions are:
\begin{enumerate}
    \item We propose the Decoupled Action Expert and use it to show that, in modulation-conditioned diffusion policies, downstream task adaptation can be carried entirely by the conditioning pathway, with a frozen, observation-free backbone shared across tasks.
    \item We demonstrate that a 5M-parameter MLP backbone matches or exceeds the 244M U-Net and the transformer on these benchmarks, indicating the action backbone is heavily over-parameterized for the tasks studied here.
    \item We show through ablation that the Stage~1 conditioning signal has little effect while skipping pretraining fails entirely: the backbone need only supply a general trajectory prior, not a specific one.
    \item We show that the frozen backbone transfers directly from external observation-free trajectory datasets without task-specific fine-tuning.
    \item We identify modulation-based conditioning as the mechanism that enables decoupling, and explain why attention-based conditioning fails in this setting. We further find that a small, weakly-structured MLP is better matched to action generation than the CNN and transformer backbones inherited from other domains (Section~\ref{sec:backbone}).
\end{enumerate}

The contribution is not the accuracy but what achieves it: the backbone can be frozen, shared across tasks, and shrunk from 244M to 5M---and, needing only kinematics, \emph{pretrained on large trajectory data (e.g., DROID~\cite{khazatsky2024droid}) that standard Diffusion Policy cannot use}, and in principle even on data generated synthetically via forward kinematics.


\section{RELATED WORK}

Recent VLA systems predominantly generate actions through one of two paradigms: autoregressive decoding (e.g., RT-2~\cite{RT-2}, OpenVLA~\cite{openvla}), or dedicated action experts that generate continuous actions via diffusion (e.g., DexVLA~\cite{DexVLA}) or flow matching (e.g., $\pi_0$~\cite{pi0}).
This paper focuses on the second paradigm. Section~\ref{sec:related_action_experts} reviews the architectures and scale of existing action experts, and Section~\ref{sec:cond_related} discusses the conditioning mechanisms used to steer them.

\subsection{Action Expert Architectures}
\label{sec:related_action_experts}

\textbf{The $\pi$ family and follow-up works.}
The most widely adopted configuration is the ${\sim}$300M-parameter flow-matching action expert introduced by $\pi_0$~\cite{pi0}, which receives VLM features via shared self-attention and generates actions through iterative denoising.
This design has been adopted by numerous subsequent works~\cite{pi0.5, OneTwoVLA, Interleave-VLA, Evo-0, ACoT-VLA, VLAW}.
Across these iterations, what changed was the conditioning interface (for example, $\pi_0.5$~\cite{pi0.5} introduced adaRMSNorm for timestep injection while retaining the shared self-attention design), but the expert's fundamental scale and role have remained fixed.

\textbf{Expert scale.}
Expert capacity spans two orders of magnitude.
On the compact end, CogACT~\cite{CogACT} attaches a DiT~\cite{DiT} action head (13--308M) to a 7B Prismatic~\cite{Prismatic-VLM} VLM, while SmolVLA~\cite{SmolVLA} and TinyVLA~\cite{TinyVLA} allocate ${\sim}$100M parameters despite aggressively compressing the VLM.
At the larger end, $\pi_0.6$~\cite{pi0.6} and GR00T-N1~\cite{GR00T-N1} scale to ${\sim}$860M alongside Gemma~3~\cite{Gemma3} 4B and Eagle-2 1.3B VLMs respectively, while DexVLA~\cite{DexVLA}, PointVLA~\cite{PointVLA}, GR-3~\cite{GR-3}, and HiMoE-VLA~\cite{HiMoE-VLA} reach ${\sim}$1B paired with 2--3B backbones~\cite{Qwen2-VL}.
CogACT's scaling studies suggest that larger action modules yield disproportionate gains~\cite{CogACT}, which may partly explain why even \emph{efficient} VLA designs preserve substantial action experts.

\begin{table}[t]
\centering
\caption{\small Action expert architectures and conditioning mechanisms in recent diffusion/flow-matching policy models. Expert capacity ranges from ${\sim}$100M in efficient VLAs~\cite{SmolVLA, TinyVLA} to over 1B in DexVLA~\cite{DexVLA}, yet whether such capacity is necessary remains unexplored.}
\label{tab:action_experts}
\footnotesize
\resizebox{\columnwidth}{!}{
\begin{tabular}{@{}lcccc@{}}
\toprule
\textbf{Model} & \textbf{Arch} & \textbf{Params} & \textbf{VLM} & \textbf{Conditioning} \\
\midrule
\multicolumn{5}{@{}l}{\textit{Standalone (no VLM):}} \\
DP-C~\cite{DP}                         & CNN U-Net     & 244M          & ---           & FiLM \\
RDT-1B~\cite{RDT-1B}                   & DiT           & 1.2B          & ---           & Cross-Attn \\
\midrule
\multicolumn{5}{@{}l}{\textit{${\sim}$100M experts:}} \\
CogACT~\cite{CogACT}                   & DiT           & 13--308M      & 7B            & Tok-Cat \\
SmolVLA~\cite{SmolVLA}                  & Transformer   & ${\sim}$100M  & 350M          & Cross-Attn \\
TinyVLA~\cite{TinyVLA}                  & Diffusion     & ${\sim}$100M  & 0.4--1.3B     & FiLM \\
\midrule
\multicolumn{5}{@{}l}{\textit{${\sim}$300M experts ($\pi_0$ family):}} \\
$\pi_0$~\cite{pi0}                     & Transformer   & 300M          & 3B            & Shared Self-Attn (SSA) \\
$\pi_0.5$~\cite{pi0.5}                  & Transformer   & 300M          & 3B            & SSA + adaRMSNorm \\
7+ follow-ups                           & Transformer   & 300M          & 3--7B         & Varies \\
\midrule
\multicolumn{5}{@{}l}{\textit{${\sim}$860M--1B experts:}} \\
$\pi_0.6$~\cite{pi0.6}                  & Transformer   & 860M          & 4B            & SSA + adaRMSNorm \\
GR00T-N1~\cite{GR00T-N1}               & DiT           & ${\sim}$860M  & 1.3B          & Cross-Attn + AdaLN \\
DexVLA~\cite{DexVLA}                    & ScaleDP       & 1B            & 2B            & FiLM + AdaLN \\
\bottomrule
\end{tabular}
}
\end{table}

Table~\ref{tab:action_experts} summarizes representative systems. Despite this architectural diversity, all dedicated action experts share the same functional role: a denoising backbone steered by a conditioning signal.
Expert capacity has scaled from ${\sim}$100M to over 1B parameters, yet whether such capacity is necessary, whether these backbones require task-specific training, and whether their inherited architecture fits the action target, has not been systematically investigated.
This paper provides the first such analysis.

\subsection{Conditioning Mechanisms}
\label{sec:cond_related}
\begin{figure*}[t]
    \centering
    \includegraphics[width=1\linewidth]{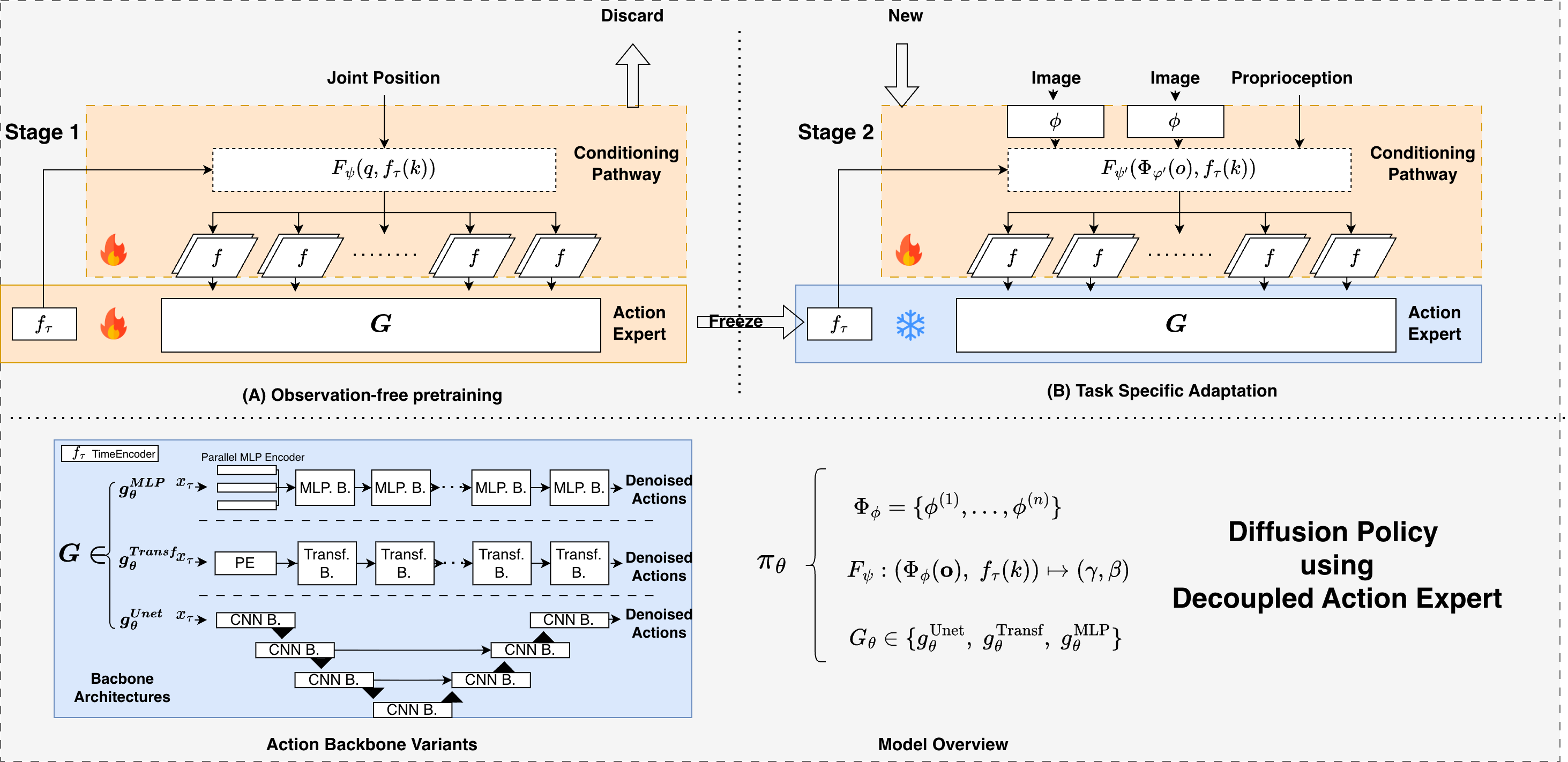}
    \caption{\small Overview of Decoupled Action Expert applied to Diffusion Policy. \textbf{(A)}~Stage~1 pretrains the backbone~$G_\theta$ and conditioning network~$F_\psi$ on observation-free forward-kinematics data. \textbf{(B)}~Stage~2 freezes~$G_\theta$ and trains new observation encoders~$\Phi_{\phi'}$ and conditioning network~$F_{\psi'}$ on task demonstrations. Bottom left: the three backbone architectures evaluated (U-Net, Transformer, MLP). Bottom right: model overview showing the policy decomposition.}
    \label{fig:DP}
\end{figure*}

Action experts receive observation and language features through a conditioning mechanism.
Existing designs fall into two structural categories that differ in where condition-specific computation resides relative to the backbone's weights.

\textbf{Attention-based conditioning} is dominant and takes three forms.
Shared self-attention, pioneered by $\pi_0$~\cite{pi0} and widely adopted~\cite{OneTwoVLA, Interleave-VLA, Evo-0, GraphCoT-VLA, Vlaser, XR-1, LingBot-VLA, GigaBrain-0, VLAW}, places VLM and action tokens in a single sequence; cross-attention~\cite{GR00T-N1, RDT-1B, GR-3, FLOWER, MemoryVLA, HiMoE-VLA, ACoT-VLA, GraspVLA} instead processes VLM features as external keys and values; and token concatenation~\cite{CogACT, Octo, BayesianVLA, VLA-JEPA} prepends condition tokens to the action sequence.
Despite their differences, all three route conditions through the backbone's own projections ($W_q$, $W_k$, $W_v$), coupling internal representations to the conditioning distribution seen during training.

\textbf{Feature modulation} takes a structurally different approach.
FiLM~\cite{FiLM} produces per-layer scale and shift parameters $(\gamma, \beta)$ via a separate conditioning network, applying element-wise affine transforms to backbone features; AdaLN~\cite{DiT} and adaRMSNorm~\cite{pi0.5} follow the same principle through learned normalization statistics.
In all cases, the backbone weights are condition-independent: different inputs steer the network through different operating regions via the external modulation signal.
This approach is used by $\pi_0.5$~\cite{pi0.5}, $\pi_0.6$~\cite{pi0.6}, GR00T-N1~\cite{GR00T-N1}, DexVLA~\cite{DexVLA}, and others~\cite{TinyVLA, CoA-VLA, Diffusion-VLA}.

The critical distinction is that modulation keeps all condition-specific computation in an external pathway, whereas attention-based methods distribute it across projection matrices inside the backbone.
We analyze how this structural difference affects backbone reuse in Section~\ref{sec:cond_ablation}.

\section{METHOD}
\label{method}

If task adaptation can be carried by the conditioning pathway alone, then the action backbone should be (1)~freezable without losing task performance and (2)~replaceable with a much smaller network.
We test both claims: the \emph{Decoupled Action Expert} recipe (Section~\ref{sec:dae}) isolates the backbone by pretraining and freezing it, and \emph{DP-MLP} (Section~\ref{sec:dp_mlp}) replaces the 244M U-Net with a 5M-parameter alternative.

\subsection{Decoupled Action Expert}
\label{sec:dae}
We decompose a FiLM-conditioned Diffusion Policy into three component groups (Fig.~\ref{fig:DP}):
observation encoders~$\Phi_\phi$, a conditioning network~$F_\psi$ that produces per-layer FiLM parameters, and a denoising backbone~$G_\theta$.
The noise prediction at diffusion step~$k$ can be written as
\begin{equation}
    \hat{\epsilon} \;=\; G_\theta\!\Big(\mathbf{a}_k,\;\; F_\psi\!\big(\Phi_\phi(\mathbf{o}),\, f_\tau(k)\big)\Big),
    \label{eq:dp_forward}
\end{equation}
where $\mathbf{o}$ denotes observations, $\mathbf{a}_k$ the noisy action sequence at diffusion step~$k$, and $f_\tau$ is a learned projection that maps the diffusion timestep to a 256-dimensional embedding.
$F_\psi$ concatenates this embedding with the observation features and outputs the scale and shift parameters $(\gamma, \beta)$ that modulate $G_\theta$ via FiLM.
The choice of FiLM conditioning is critical: because modulation keeps all condition-specific computation in the external pathway~$F_\psi$, the backbone~$G_\theta$ remains condition-independent and can be frozen without losing the ability to represent new tasks (Section~\ref{sec:cond_related}).
In standard training, all parameters $\{\phi, \psi, \theta\}$ are optimized jointly on observation--action demonstrations.
Our decoupled recipe separates this into two stages.

\subsubsection{Stage 1: Observation-Free Pretraining}
We exploit the deterministic forward-kinematics mapping from joint positions~$\mathbf{q}$ to end-effector poses~$\mathbf{p}$ to construct observation-free training pairs (Fig.~\ref{fig:DP}, Stage~1).
The backbone and conditioning network are trained to denoise end-effector pose action sequences conditioned on joint positions:
\begin{equation}
    \hat{\epsilon} \;=\; G_\theta\!\Big(\mathbf{a}_k,\;\; F_\psi\!\big(\mathbf{q},\, f_\tau(k)\big)\Big).
    \label{eq:stage1}
\end{equation}
Trainable parameters: $\{\psi, \theta\}$.
No observation encoder is needed: $\mathbf{q}$ enters the conditioning network directly.
The training pairs~$(\mathbf{q}, \mathbf{p})$ can be extracted from any existing trajectory dataset at negligible cost, since joint positions and end-effector poses are standard proprioceptive signals, or generated synthetically via forward kinematics without any environment interaction.

\subsubsection{Stage 2: Task-Specific Adaptation}
The pretrained backbone $G_{\bar\theta}$ is frozen (Fig.~\ref{fig:DP}, Stage~2).
New observation encoders $\Phi_{\phi'}$ and a new conditioning network $F_{\psi'}$ are initialized and trained on task demonstrations:
\begin{equation}
    \hat{\epsilon} \;=\; G_{\bar\theta}\!\Big(\mathbf{a}_k,\;\; F_{\psi'}\!\big(\Phi_{\phi'}(\mathbf{o}),\, f_\tau(k)\big)\Big).
    \label{eq:stage2}
\end{equation}
Trainable parameters: $\{\phi', \psi'\}$. Frozen: $\{\bar\theta\}$.
The observation encoder~$\Phi_{\phi'}$ is modular and can be substituted with any pretrained encoder or extended to other modalities.

\subsection{Lightweight Backbone: DP-MLP}
\label{sec:dp_mlp}
The backbone~$G_\theta$ in DP-C is a 244M-parameter 1D convolutional U-Net~\cite{DP}.
A manipulation policy generates only $16 \times 10 = 160$ action values in DP, suggesting this capacity may be excessive for such low-dimensional targets.
We propose DP-MLP, which replaces the U-Net with $L$ residual FiLM-MLP blocks totaling approximately 5M parameters (${\sim}50\times$ reduction).
Each horizon step is embedded via a per-position linear projection, and each block applies FiLM modulation from~$F_\psi$ between two linear layers with a residual connection and layer normalization:
\begin{equation}
    \mathbf{h}^{(\ell+1)} = \text{LN}\!\big(\mathbf{h}^{(\ell)} + W_2^{(\ell)}(\boldsymbol{\gamma}^{(\ell)} \odot \text{GELU}(W_1^{(\ell)}\, \mathbf{h}^{(\ell)}) + \boldsymbol{\beta}^{(\ell)})\big).
    \label{eq:film_mlp}
\end{equation}
Because DP-MLP uses FiLM conditioning, it is directly compatible with the decoupled recipe described above. We evaluate it against the 244M U-Net in Section~\ref{sec:lightweight}.

\section{EXPERIMENTS}
\label{sec:experiments}

\label{sec:setup}

\textbf{Benchmarks.}
We evaluate on two standard manipulation benchmarks: MimicGen~\cite{mimicgen} and LIBERO~\cite{liu2023libero}.
\textit{MimicGen} extends the \textit{robomimic} benchmark with a large set of automatically generated manipulation tasks.
We select eight tasks that span coarse manipulation, fine-grained manipulation, and multi-step manipulation, each with 1000 demonstrations, and train a separate model per task.
\textit{LIBERO}~\cite{liu2023libero} provides four task suites (Spatial, Object, Goal, and Long), each containing 10 language-conditioned manipulation tasks with 50 demonstrations per task.
We train one model per suite.
This single-task (MimicGen) and per-suite (LIBERO) protocol follows EquiDiff~\cite{equidiff} and OpenVLA~\cite{openvla}.

\textbf{Implementation.}
On MimicGen, we adopt the original Diffusion Policy implementation~\cite{DP}.
Observations consist of front-view and wrist-view camera images encoded by ResNet-18, concatenated with proprioceptive state (end-effector position, quaternion orientation, and gripper openness).
On LIBERO, we adopt the language-conditioned Diffusion Policy from DROID~\cite{khazatsky2024droid}, which is the standard baseline used by OpenVLA~\cite{openvla} and subsequent works.
Observations consist of front-view and wrist-view camera images encoded by ImageNet-pretrained ResNet-50, concatenated with proprioceptive state and a frozen DistilBERT~\cite{sanh2019distilbert} language embedding of the task instruction.
All observation features are concatenated into a single conditioning vector that modulates the action backbone via FiLM.
The DP-MLP variant is described in Section~\ref{sec:dp_mlp}.
Notably, in the decoupled setting, the language embedding, like all other observation features, enters through the conditioning pathway trained in Stage~2, while the frozen backbone remains identical to the observation-free Stage~1 pretraining.
Actions are 10-dimensional (3 position + 6 rotation in continuous 6D representation~\cite{zhou2019continuity} + 1 gripper) with a prediction horizon of 16 steps.

\textbf{Evaluation.}
Unless otherwise noted, experiments are run with three random seeds (0, 42, 420) and we report mean $\pm$ standard deviation. The conditioning mechanism ablation (Section~\ref{sec:cond_ablation}) uses a single seed (42).
On MimicGen, we evaluate every checkpoint throughout training and report the maximum success rate, following the common practice in single-task manipulation policy evaluation~\cite{equidiff}.
On LIBERO, following OpenVLA~\cite{openvla}, we evaluate only the final checkpoint; each seed runs 500 rollouts per suite (50 per task $\times$ 10 tasks).

\subsection{Does Decoupling Preserve Task Performance?}
\label{sec:decoupling}

The central claim of this paper is that, in modulation-conditioned diffusion policies, downstream task adaptation can be carried entirely by the conditioning pathway, leaving the action backbone a frozen, task-supervision-free module.
We test this by applying the same decoupled procedure (pretrain the action backbone on observation-free kinematic data, freeze it entirely, and train only the conditioning pathway on downstream tasks) to two diffusion policy variants that differ in backbone architecture and conditioning mechanism: DP-C (U-Net, FiLM) and DP-T (transformer, cross-attention).
If performance is preserved, the frozen backbone required no task-specific training and task adaptation was carried by the conditioning pathway alone.
Results are shown in Table~\ref{tab:main_results} (DP-MLP is discussed in Section~\ref{sec:lightweight}).

\textbf{Scope of the claim.} We frame this result operationally. We show that task adaptation \emph{can} be carried entirely by the conditioning pathway, leaving the backbone frozen and free of task-specific data; we do \emph{not} claim the backbone is representationally task-agnostic or performs little of the task-relevant computation. The frozen backbone is in fact necessary---a randomly initialized one fails entirely (Section~\ref{sec:ablation})---and supplies general, embodiment-level trajectory structure that is load-bearing. Whether the bulk of task-relevant computation resides in the frozen backbone or in the expressive conditioning pathway---structural confinement vs.\ conditioning compensation---is not identified by task performance alone, and we leave it to future work. Our capacity claims (Section~\ref{sec:lightweight}) concern these benchmarks; we use Diffusion Policy as a controlled proxy and do not test VLA systems.

\begin{table*}[t]
\centering
\caption{\small Normal vs.\ decoupled training across three architectures on MimicGen (8 tasks) and LIBERO (4 suites). FiLM-conditioned architectures DP-C (244M U-Net) and DP-MLP (5M MLP) preserve performance under decoupling, while DP-T (transformer, cross-attention) collapses (analyzed in Section~\ref{sec:cond_ablation}). The preserved performance of a frozen, task-supervision-free backbone shows that task adaptation can be carried entirely by the conditioning pathway. The 5M MLP matching or exceeding the 244M U-Net on both benchmarks further indicates the action backbone is over-parameterized for these tasks.}
\label{tab:main_results}
\setlength{\tabcolsep}{4pt}
\small
\renewcommand{\arraystretch}{0.5}
\textit{MimicGen:}
\vspace{2pt}

\begin{tabular*}{\textwidth}{@{\extracolsep{\fill}}llccccccccc@{}}
\toprule
              &           &  Stack(A) &  Square(B) &  Coffee(C) &  Thread(D) &  Stack3(E) &  Hammer(F) &  3Piece(G) &  Mug(H) &  Avg \\
\midrule
DP-C          & Normal    & 100$\pm$0 & 55$\pm$5 & 73$\pm$2 & 47$\pm$7 & 95$\pm$2 & 54$\pm$5 & 17$\pm$2 & 67$\pm$1 & 63.6$\pm$1.8 \\
              & Decoupled & 100$\pm$0 & 50$\pm$2 & 71$\pm$4 & 39$\pm$2 & 86$\pm$6 & 58$\pm$5 & 25$\pm$3 & 69$\pm$1 & \textbf{62.2$\pm$2.3} \\
\midrule
DP-T          & Normal    &  97$\pm$1 & 41$\pm$2 & 74$\pm$6 & 29$\pm$2 & 83$\pm$7 & 65$\pm$2 & 29$\pm$3 & 66$\pm$2 & 60.4$\pm$0.8 \\
              & Decoupled &  62$\pm$2 &  5$\pm$1 &  2$\pm$0 &  5$\pm$2 & 25$\pm$3 & 28$\pm$7 &  3$\pm$1 & 25$\pm$4 & 19.4$\pm$0.4 \\
\midrule
DP-MLP        & Normal    & 100$\pm$0 & 51$\pm$1 & 76$\pm$4 & 47$\pm$2 & 89$\pm$1 & 61$\pm$3 & 36$\pm$3 & 69$\pm$1 & \textbf{65.9$\pm$0.7} \\
              & Decoupled & 100$\pm$0 & 41$\pm$3 & 63$\pm$1 & 41$\pm$6 & 77$\pm$2 & 63$\pm$3 & 33$\pm$3 & 73$\pm$5 & 61.2$\pm$0.5 \\
\bottomrule
\end{tabular*}

\vspace{6pt}
\textit{LIBERO:}
\vspace{2pt}

\begin{tabular*}{\textwidth}{@{\extracolsep{\fill}}llccccc@{}}
\toprule
              &           & Spatial & Object & Goal & Long & Avg \\
\midrule
DP-C          & Normal    & 89.5$\pm$1.2 & 93.1$\pm$2.8 & 78.3$\pm$0.6 & 56.2$\pm$1.8 & 79.3$\pm$0.6 \\
              & Decoupled & 87.7$\pm$1.1 & 92.8$\pm$1.2 & 74.5$\pm$1.2 & 52.3$\pm$2.0 & 76.8$\pm$0.9 \\
\midrule
DP-T          & Normal    & 75.5$\pm$0.3 & 96.6$\pm$0.4 & 77.9$\pm$1.0 & 55.5$\pm$0.4 & 76.4$\pm$0.4 \\
              & Decoupled & 4.2$\pm$1.2 & 8.6$\pm$1.1 & 8.0$\pm$0.9 & 2.8$\pm$0.4 & 5.9$\pm$0.2 \\
\midrule
DP-MLP        & Normal    & 92.0$\pm$1.6 & 97.7$\pm$0.1 & 82.1$\pm$0.6 & 67.1$\pm$0.8 & \textbf{84.7$\pm$0.5} \\
              & Decoupled & 87.1$\pm$1.2 & 98.6$\pm$0.3 & 82.0$\pm$1.2 & 68.9$\pm$1.6 & \textbf{84.2$\pm$0.9} \\
\bottomrule
\end{tabular*}
\end{table*}

\textbf{DP-C decoupling preserves performance.}
Recall what decoupling entails: the backbone is pretrained entirely on observation-free kinematic pairs (it has never seen a single task image) and is then frozen.
Only the conditioning pathway is trained on downstream demonstrations.
On MimicGen, this frozen, task-supervision-free backbone achieves 62.2$\pm$2.3\% compared to 63.6$\pm$1.8\% under normal training, a drop of only 1.4 points.
Per-task results are mixed: decoupling slightly degrades some tasks (Stack3 $-$9, Thread $-$8) while improving others (3Piece $+$8, Hammer $+$4), with confidence intervals largely overlapping.
On LIBERO, the gap is slightly larger ($-$2.5: 76.8$\pm$0.9\% vs.\ 79.3$\pm$0.6\%), concentrated in the harder Goal ($-$3.8) and Long ($-$3.9) suites while Spatial and Object remain within noise.
This may partially stem from having fewer trainable parameters under decoupling, since the frozen backbone leaves only the conditioning layers updatable.
Overall, that a frozen, task-supervision-free backbone performs comparably to one trained end-to-end shows that task adaptation can be carried by the conditioning pathway without updating the backbone.

\textbf{Conditioning mechanism determines compatibility.}
Not all conditioning mechanisms survive decoupling, however.
DP-T, which injects observations via cross-attention, collapses from 60.4$\pm$0.8\% to 19.4$\pm$0.4\% on MimicGen and from 76.4$\pm$0.4\% to 5.9$\pm$0.2\% on LIBERO, near-total failure despite following the same decoupled procedure as DP-C.
Since DP-C and DP-T differ in both backbone architecture and conditioning mechanism, this comparison alone cannot isolate the cause.
We separate these two factors in Section~\ref{sec:cond_ablation} by testing multiple conditioning mechanisms on the same backbone.

\subsection{Lightweight Action Backbone}
\label{sec:lightweight}

As discussed in Section~\ref{sec:related_action_experts}, action experts in recent VLA models typically range from ${\sim}$100M to over 1B parameters, following design conventions from image diffusion models that denoise high-dimensional structured outputs: Stable Diffusion~\cite{stable_diffusion} produces $64{\times}64{\times}4{=}16{,}384$ latent values, SDXL~\cite{podell2023sdxl} $128{\times}128{\times}4{=}65{,}536$, and FLUX~\cite{labs2025flux} $128{\times}128{\times}16{=}262{,}144$.
A manipulation policy, by contrast, generates only short sequences of physically correlated, low-dimensional actions ($16 \times 10 = 160$ values in DP, or $50 \times 18 = 900$ in $\pi_0$~\cite{pi0}), two to three orders of magnitude smaller.
This gap suggests that current action backbones may be heavily over-parameterized for such low-dimensional action targets.
We test this using DP-MLP (Section~\ref{sec:dp_mlp}), a 5M-parameter backbone that represents a ${\sim}50\times$ reduction from the U-Net.
Results are shown in Table~\ref{tab:main_results}.

\textbf{A 5M MLP exceeds a 244M U-Net.}
Under normal training, DP-MLP achieves 65.9$\pm$0.7\% on MimicGen and 84.7$\pm$0.5\% on LIBERO, both exceeding DP-C (63.6$\pm$1.8\% and 79.3$\pm$0.6\%) despite a ${\sim}50\times$ parameter reduction.
Per-task gains are especially pronounced on 3Piece ($+$19) and Hammer ($+$7) in MimicGen, and on Long ($+$10.9) in LIBERO, while only Stack3 ($-$6) trails noticeably.
Under decoupling, DP-MLP retains 61.2$\pm$0.5\% on MimicGen and 84.2$\pm$0.9\% on LIBERO.
The MimicGen decoupled drop ($-$4.7) is larger than DP-C's ($-$1.4), driven by Coffee ($-$13), Stack3 ($-$12), and Square ($-$10). On LIBERO the drop is negligible ($-$0.5), with Goal and Long essentially unchanged and only Spatial declining ($-$4.9), a different pattern from DP-C, where the loss concentrated in Goal and Long.
Overall, the normal training results show that a 5M MLP can exceed a 244M U-Net, while the decoupled training results further show that the action backbone needs no task-specific training.

\subsection{Pretraining on External Trajectory Datasets}
\label{sec:ood}

Because Stage~1 pretraining requires only observation-free kinematics, the decoupled backbone can be pretrained on \emph{any} trajectory dataset containing joint-state information, a data source that standard Diffusion Policy training cannot leverage at all, since it requires paired observation--action data.
We test this with DROID~\cite{khazatsky2024droid}, a large-scale dataset of ${\sim}$76k real-world Franka manipulation trajectories whose tasks and environments share no overlap with MimicGen or LIBERO.
We extract only joint positions and end-effector poses, discarding all images and task labels.
This also constitutes a natural data scaling experiment: in-distribution FK data is limited to the target benchmark's own demonstrations (8{,}000 for MimicGen, 2{,}000 for LIBERO), whereas DROID contributes an order of magnitude more trajectories.
Table~\ref{tab:ood} compares two pretraining sources: in-distribution FK and DROID FK.

\begin{table}[t]
\centering
\caption{\small Effect of pretraining data source on DP-C Decoupled. $\Delta_\text{N}$: difference from Normal, $\Delta_\text{ID}$: difference from In-Dist FK. DROID provides ${\sim}$76k external Franka trajectories whose observation-free kinematics are inaccessible to standard DP training. MimicGen: mean $\pm$ std over 3 seeds (max checkpoint). LIBERO: mean $\pm$ std over 3 eval seeds (single training seed).}
\label{tab:ood}
\small
\resizebox{\columnwidth}{!}{
\begin{tabular}{@{}lcccccc@{}}
\toprule
                    & \multicolumn{3}{c}{\textit{MimicGen}} & \multicolumn{3}{c}{\textit{LIBERO}} \\
\cmidrule(lr){2-4} \cmidrule(lr){5-7}
\textbf{Pretrain}   & Avg           & $\Delta_\text{N}$  & $\Delta_\text{ID}$ & Avg           & $\Delta_\text{N}$  & $\Delta_\text{ID}$ \\
\midrule
Normal              & 63.6$\pm$1.8 & --   & --   & 79.3$\pm$0.6  & --   & -- \\
\midrule
Dec (In-Dist FK)    & 62.2$\pm$2.3  & $-$1.4 & --   & 76.8$\pm$0.9  & $-$2.5 & -- \\
Dec (DROID FK)      & 63.8$\pm$1.2  & +0.2   & +1.6 & 78.3$\pm$0.4  & $-$1.0 & +1.5 \\
\bottomrule
\end{tabular}
}
\end{table}

DROID-pretrained backbones outperform in-distribution pretraining on both benchmarks ($\Delta_\text{ID}$: $+$1.6 on MimicGen, $+$1.5 on LIBERO), despite the complete distribution shift.
This suggests that the backbone learns general kinematic structure rather than task-specific patterns, and that larger, more diverse trajectory collections can directly improve the decoupled recipe, a data source entirely inaccessible to standard end-to-end training.
We note that our Stage~1 pretraining is a straightforward implementation without careful design of learning rate schedules, data mixing, or curriculum. These experiments serve as an illustration that external trajectory pretraining can benefit decoupled performance, and we expect the actual gains to grow with more deliberate engineering.
\subsection{Ablation: Stage~1 Conditioning}
\label{sec:ablation}

Having established that a frozen, observation-free backbone preserves task performance (Section~\ref{sec:decoupling}), we now ask: what does the backbone actually learn during Stage~1, and does the choice of conditioning signal matter?
We design four variants that share an identical Stage~2 procedure (freeze backbone, train new FiLM layers on MimicGen) but differ only in Stage~1.
\textbf{JP conditioned} (JP $\to$ eePose) is the standard method from Section~\ref{sec:decoupling}: the backbone denoises end-effector trajectories conditioned on joint positions via forward kinematics.
\textbf{Unconditional} ($\varnothing \to$ eePose) removes all conditioning, training the backbone to denoise end-effector trajectories without any input signal.
\textbf{eePose self-conditioned} (eePose $\to$ eePose) conditions on the same end-effector trajectories, creating a trivial identity mapping.
\textbf{Random frozen} skips Stage~1 entirely, freezing a randomly initialized backbone.
Results are shown in Table~\ref{tab:ablation}.

\begin{table}[t]
\centering
\caption{\small Stage~1 conditioning ablation on MimicGen (DP-C Decoupled). All variants share the same Stage~2 (freeze backbone, train new FiLM layers). Random frozen skips Stage~1 entirely. Mean $\pm$ std over 3 seeds (max checkpoint). Task letters A--H correspond to those in Table~\ref{tab:main_results}.}
\label{tab:ablation}
\small
\resizebox{\columnwidth}{!}{
\begin{tabular}{@{}lcccc@{}}
\toprule
\textbf{Variant}                                 &  A &  B &  C &  D \\
\midrule
Random frozen                                    & 0$\pm$0 & 0$\pm$0 & 0$\pm$0 & 0$\pm$0 \\
Unconditional ($\varnothing \to$ eePose)         & 99$\pm$1 & 53$\pm$1 & 72$\pm$0 & 39$\pm$2 \\
eePose self-cond (eePose $\to$ eePose)           & 99$\pm$1 & 51$\pm$5 & 74$\pm$0 & 41$\pm$3 \\
JP conditioned (JP $\to$ eePose)                 & 100$\pm$0 & 50$\pm$2 & 71$\pm$4 & 39$\pm$2 \\
\bottomrule
\end{tabular}
}

\vspace{4pt}

\resizebox{\columnwidth}{!}{
\begin{tabular}{@{}lcccccc@{}}
\toprule
\textbf{Variant}                                 &  E &  F &  G &  H & Avg \\
\midrule
Random frozen                                    & 0$\pm$0 & 0$\pm$0 & 0$\pm$0 & 0$\pm$0 & 0.0$\pm$0.0 \\
Unconditional                                    & 85$\pm$3 & 59$\pm$3 & 27$\pm$1 & 64$\pm$2 & 62.2$\pm$0.5 \\
eePose self-cond                                 & 82$\pm$2 & 61$\pm$1 & 32$\pm$6 & 69$\pm$1 & 63.8$\pm$1.1 \\
JP conditioned                                   & 86$\pm$6 & 58$\pm$5 & 25$\pm$3 & 69$\pm$1 & 62.2$\pm$2.3 \\
\bottomrule
\end{tabular}
}
\end{table}

\textbf{Pretraining is essential, but the conditioning signal is not.}
The randomly initialized backbone achieves 0\% across all tasks, confirming that Stage~1 pretraining is necessary for the frozen backbone to be useful.
However, the three pretrained variants achieve comparable average success rates (62.2$\pm$0.5\%, 63.8$\pm$1.1\%, 62.2$\pm$2.3\%), with overlapping confidence intervals.
Since Stage~1 FiLM layers are discarded and replaced in Stage~2, only the U-Net weights carry over, and these weights provide a downstream-equivalent trajectory prior whether pretrained on joint positions, end-effector poses, or nothing at all.
The backbone therefore serves as a general trajectory prior (temporal correlations, smooth motion priors, and end-effector geometry) whose downstream usefulness does not depend on the specific Stage~1 conditioning signal.
This simplifies the practical requirements of the decoupled recipe: Stage~1 pretraining can use any available trajectory data without requiring paired kinematic inputs.

\subsection{Ablation: Conditioning Mechanism}
\label{sec:cond_ablation}

\begin{figure}[t]
\centering
\includegraphics[width=\columnwidth]{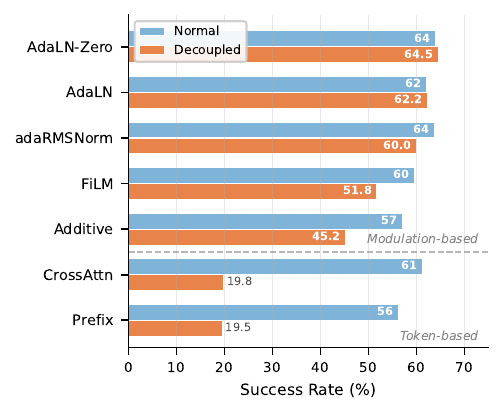}
\caption{\small Conditioning mechanism ablation on DP-T (MimicGen, seed~42). Seven mechanisms on the same transformer backbone. Under normal training all perform comparably (56--64\%). Under decoupling, modulation-based methods (top) preserve performance while token-based methods (bottom) collapse.}
\label{tab:cond_ablation}
\end{figure}

As noted in Section~\ref{sec:decoupling}, DP-C (FiLM) preserves performance under decoupling while DP-T (cross-attention) collapses, but the two architectures differ in both backbone and conditioning, leaving the root cause ambiguous.
To isolate the role of the conditioning mechanism, we evaluate seven methods on the same transformer backbone ($d{=}256$, 8 layers): Cross-Attention, Prefix tuning~\cite{li2021prefix}, FiLM~\cite{FiLM}, AdaLN, AdaLN-Zero~\cite{DiT}, adaRMSNorm~\cite{pi0.5}, and Additive injection.
Results under normal and decoupled training (seed~42) are shown in Fig.~\ref{tab:cond_ablation}.

Under normal training, all seven mechanisms achieve comparable performance (56--64\%), confirming that the conditioning choice does not inherently limit the backbone.
Under decoupling, a clear pattern emerges based on how conditioning interacts with backbone computation.
Token-based methods that embed condition-specific projections inside the backbone collapse: Cross-Attention drops from 61.3 to 19.8\% and Prefix from 56.3 to 19.5\%, because freezing the backbone fixes these projections to the Stage~1 input distribution.
Modulation-based methods show a gradient of robustness: AdaLN-Zero and AdaLN fully preserve performance (64.0 $\to$ 64.5\% and 62.0 $\to$ 62.3\%), adaRMSNorm retains most (63.8 $\to$ 60.0\%), FiLM degrades moderately (59.5 $\to$ 51.8\%), and Additive degrades further (57.0 $\to$ 45.3\%).
The determining factor is whether conditioning computation \textbf{separates from backbone parameters}: modulation-based methods that keep conditioning external to the backbone are robust to freezing, while token-based methods that route conditions through backbone projections collapse.
Notably, several recent VLA models have independently adopted modulation-based conditioning ($\pi_0.5$ uses adaRMSNorm, GR00T-N1~\cite{GR00T-N1} uses AdaLN, DexVLA~\cite{DexVLA} uses FiLM), consistent with our finding.
All results in this ablation use a single fixed seed (42).

\subsection{Which Backbone Fits Action Generation?}
\label{sec:backbone}

Action experts inherit their backbones, and the inductive biases that come with them, from image, video, and language generation. But the action target is a different object: in our Diffusion Policy setup, a short sequence of $16 \times 10 = 160$ physically coupled values, replanned after a few steps---orders of magnitude smaller than the high-dimensional signals those architectures were built for. We ask which backbone actually fits this target, comparing the three architectures.

\textbf{The 5M MLP performs best in our experiments.} Under normal training, the 5M MLP is the best backbone in Table~\ref{tab:main_results} (Section~\ref{sec:lightweight}), ahead of the 244M CNN U-Net on both MimicGen and LIBERO while using ${\sim}50\times$ fewer parameters. The two configurations share the same FiLM conditioning and differ only in their backbone. Here the smaller backbone is the stronger one.

\textbf{The gap suggests an inductive-bias mismatch.} A U-Net's downsampling hierarchy and a Transformer's long-range attention are designed for high-dimensional, structured signals. On a short, low-dimensional trajectory the downsampling has little to compress and the attention spans a near-trivial sequence. The MLP imposes fewer such assumptions and learns the trajectory prior directly. The lesson is not that less inductive bias is always better, but that the backbone's bias should match the action target rather than be inherited from another domain. We offer this as an interpretation rather than a definitive ranking, since our benchmarks are short-horizon and low-diversity and we do not separately ablate downsampling, attention, and capacity.

\section{Limitations and Future Work}
\label{sec:conclusion}

This study deliberately uses Diffusion Policy as its testbed to enable the hundreds of controlled trials and systematic ablations needed to isolate the role of the action backbone.
Because VLA action experts, such as $\pi_0$'s flow-matching network and CogACT's DiT, serve the same functional role as the denoising backbone studied here, we expect the core findings to transfer, but directly validating the decoupled recipe on full VLA systems remains important future work.
All experiments are conducted in simulation (MimicGen and LIBERO), and real-world validation is needed to confirm that the decoupled training transfers to physical manipulation.
Our DROID pretraining experiments (Section~\ref{sec:ood}) use a straightforward implementation without optimized learning rate schedules, data mixing, or curriculum.
A well-designed pretraining recipe could further improve the quality of the frozen backbone, making this a promising direction for future work.
Finally, since diffusion and flow-matching models call the backbone repeatedly during inference, a smaller backbone directly reduces action generation latency.
We showed that a 5M MLP matches or exceeds a 244M U-Net in task performance, but did not explore the minimum effective backbone size or the resulting inference speedup, both of which are worth investigating for deployment.

{
\small
\bibliographystyle{plainnat}
\bibliography{refs/Datasets,refs/DPRelated,refs/misc,refs/newVLAs}
}

\end{document}